%% file: paper.tex
\documentclass{article}

\usepackage{amsmath,amsfonts,amssymb,times,graphicx,natbib,algorithm,algorithmic,hyperref}

\usepackage[accepted]{whi2016}

\icmltitlerunning{Enhancing Transparency and Control when Drawing Data-Driven Inferences about Individuals}

\begin{document}

\twocolumn[
\icmltitle{Enhancing Transparency and Control when Drawing Data-Driven Inferences about Individuals}

\icmlauthor{Daizhuo Chen}{dchen16@gsb.columbia.edu}
\icmladdress{Columbia Business School,
             2960 Broadway, New York, NY 10029 USA}
\icmlauthor{Samuel P. Fraiberger}{s.fraiberger@neu.edu}
\icmladdress{Network Science Institute at Northeastern University,
             177 Huntington Avenue, Boston, MA 02114 USA}
\icmlauthor{Robert Moakler}{rmoakler@stern.nyu.edu}
\icmlauthor{Foster Provost}{fprovost@stern.nyu.edu}
\icmladdress{NYU Stern School of Business,
             44 West 4th Street, New York, NY 10012 USA}
\vskip 0.3in
]

\begin{abstract}
Recent studies have shown that information disclosed on social network sites (such as Facebook) can be used to predict personal characteristics with surprisingly high accuracy. In this paper we examine a method to give online users transparency into why certain inferences are made about them by statistical models, and control to inhibit those inferences by hiding (``cloaking'') certain personal information from inference. We use this method to examine whether such transparency and control would be a reasonable goal by assessing how difficult it would be for users to actually inhibit inferences. Applying the method to data from a large collection of real users on Facebook, we show that a user must cloak only a small portion of her Facebook Likes in order to inhibit inferences about their personal characteristics. However, we also show that in response a firm could change its modeling of users to make cloaking more difficult.
\end{abstract} 

\section{Introduction}

Successful pricing strategies as well as marketing and political campaigns depend on the ability to target consumers or voters accurately. This generates incentives to acquire and exploit information about personal characteristics such as gender, marital status, religion, and sexual or political orientation.  Personal characteristics are often hard to observe because of lack of data or privacy restrictions. As a result, firms and governments increasingly depend on statistical inferences drawn from available information.

Online user targeting systems increasingly are trained using information on users' fine-grained behaviors~\cite{perlich2014machine}. A growing trend is to base targeting on information disclosed by users on social networks. For instance, Facebook has recently deployed a system that allows third party applications to display ads on their platform using their user's profile information, such as the things they indicate that they ``Like."\footnote{https://developers.facebook.com/blog/post/2014/10/07/audience-network}\footnote{We will capitalize ``Like'' when referring to the action or its result on Facebook.}

%A predictive model can be used to give each user a score that is proportional to the probability of having a certain personal trait, such as being gullible, introverted, female, a drug user, gay, etc.~\cite{kosinski2013private}, or the probability to respond positively to some action~\cite{perlich2014machine}. Users then can be targeted based on their predicted propensities.

%In practice, model confidence plus a budget for showing content leads to targeting users in some top quantile of the score distribution given by predictive models~\cite{perlich2014machine}.  

While some individuals may benefit from being targeted based on inferences of their personal characteristics, others may find such inferences unsettling.  These inferences may be incorrect due to a lack of data, inadequate models, or simply the probabilistic nature of the inference.  Moreover, some users may not wish to have certain personal characteristics inferred at all. To many, privacy invasions via statistical inferences are at least as troublesome as privacy invasions based on revealing personal data~\cite{BarocasThesis2014}. However, social networks such as Facebook lack features that allow for transparency and fine-grained control over how users' profiles and behavioral information are used to determine targeted content and advertisements.

%\footnote{In 2014, Facebook introduced a feature called ``Why am I seeing this ad?'' which gives users partial transparency into why they are being targeted. Users can also selectively cloak particular categories of ads or advertisers; they can also modify their ``ad preferences'' to hide categories of information from being used for targeting.
%However it does not currently allow fine-grained control over inferences, which is the topic of the paper.This recent development by Facebook may be viewed as evidence for the value of a fine-grained approach such as the one we present.}

%In response to an increase in demand for privacy from online users, suppliers of browsers such as Chrome and Firefox have developed features such as ``Do Not Track,'' ``Incognito,'' and ``Private Windows'' to control the collection of information about web browsing. However, these features provide neither clear transparency into what inferences are drawn and why, nor easy, fine-grained control over what information may be used for inference. Furthermore, 

% In this paper, we draw on an idea introduced for explaining document classifications~\cite{MartensProvostMISQ} as a means for providing transparency into the reasons why a particular inference is drawn about an individual.  
Transparency is based on understanding the reasons for data-driven inferences.  We ask: what is the minimal set of evidence such that, if it had not been present, an inference about a user would not have been drawn?  We introduce the idea of a ``cloaking device'' as a vehicle to offer users control over inferences, and we show that a user needs to only cloak a small portion of her Likes in order to inhibit inference, based on the modeling presented by~\cite{kosinski2013private}.  %Specifically, the cloaking device provides a mechanism for users to inhibit the use of particular pieces of information in inference.  Importantly, the user can cloak particular information \textit{from inference}, without having to stop sharing the information with his social network friends.

\section{Privacy, Cloakability, and the Evidence Counterfactual} 

Online privacy is becoming an increasing concern for consumers, regulators and policy makers~\cite{WhiteHouse2012}. The analytics literature has traditionally focused on the issue of privacy of personal information (see~\cite{Smith11,pavlou2011state} for an overview). However, with the rapid increase in the amount of social media data available, statistical inference about personal characteristics is a growing concern~\cite{BarocasThesis2014,Schoen13}. A series of papers have shown the predictive power of information disclosed on Facebook to infer users' personal characteristics~\cite{kosinski2013private,bachrach2012personality,schwartz2013personality}. A recent study based on a survey of Facebook users found that users do not feel that they have the appropriate tools to mitigate their privacy concerns when it comes to social network data~\cite{Johnson:2012:FPC:2335356.2335369}.

%There is evidence that when given the appropriate tools, people will give up some benefits they derive from social network activity in order to satisfy their privacy concerns~\cite{Knijnenburg13}. 
 
\begin{figure}[h]
  \centerline{\includegraphics[width=1.00\linewidth]{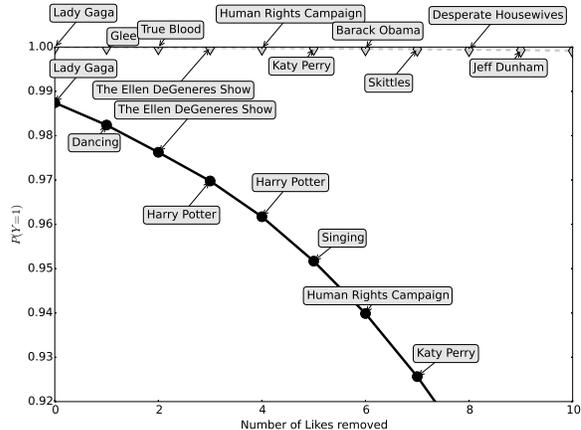}}
  \caption{The predicted probability of being gay as a function of Like cloaking for two users. For each line, the leftmost point is the estimated probability of being gay for the user before cloaking.  Moving left to right, for each user, Likes are removed one-by-one from consideration by the inference procedure in order of greatest effect on the estimated score. One user's probability drops dramatically after cloaking fewer than ten Likes; the other's is hardly affected at all.}
  \label{fig:removalline}
  \vspace*{-6mm}
\end{figure}

The method we propose to provide transparency over inferences is based on the idea of explaining individual model predictions by examining the ``evidence'' in the input feature vectors that, if removed, would lead the model not to make the prediction (introduced in the context of data-driven document classifications~\cite{MartensProvostMISQ}).  As a shorthand, we refer to this counterfactual notion of ``what would the model have done if this evidence hadn't been present'' as an ``evidence counterfactual.'' The prior work can be generalized directly to any domain where the features taken as input can be seen as interpretable pieces of evidence for or against a particular non-default inference.\footnote{The explanation for a default prediction---namely, that there is no evidence for any alternative---often will be viewed as either trivial or unsatisfying. See~\cite{MartensProvostMISQ} for further discussion and other nuances of explaining model-based inferences.}  
% Consider the increasingly common scenario~\cite{EnricBigData2014} where there is a vast number of possible pieces of evidence, but any individual only exhibits a very small number of them---such as when inferences are drawn from Likes on Facebook. We can provide transparency by applying the methods presented by~\cite{MartensProvostMISQ} to create one or more evidence counterfactual explanations for any non-default classification. 
Figure~\ref{fig:removalline} illustrates the case of two users, their probabilities of being gay as predicted by the inference procedure of~\cite{kosinski2013private}, and the effect of removing evidence from their data. As evidence is removed by cloaking Likes, we see that removing fewer than ten Likes for one user results in a dramatic drop in the predicted probability of being gay, whereas for the other user, removing the same number of Likes reduces the probability hardly at all. This motivating example leads us to propose a general method of cloaking. 

%\footnote{As with predictive modeling projects generally, engineering the right representation often is key to top-level performance.  So for example, one might code the lack of a particularly popular Like as positive evidence.  We will only consider the presence of a Like in our results, but our qualitative results should generalize across such alternative representation engineering.} 

%Given an individual, a specific model-based inference about the individual, and an evidence counterfactual explanation for why the inference was made, we can now describe the core design, use, and value of the cloaking device.  The cloaking device allows the individual to hide (to ``cloak") particular evidence, e.g., one or more Likes, from the inference procedure.  Specifically, once a Like is cloaked, the inference procedure would remove it from its input for this user, and therefore treat the user as if she had not Liked this item.  The evidence counterfactual presents the user with a minimal set of Likes to cloak in order to inhibit the inference.  

%Consider the task of predicting whether or not a user is gay using Facebook Likes. While users might choose to disclose on the platform that they are gay, some may not wish to make this fact available to advertisers or others modeling online user behavior.  A user who has not shared this status may not want it to be predicted by the system. In addition, a user who is in fact not gay may not want an incorrect inference to be drawn about him. 

%The cloaking device thus has two important dimensions of value.  First, it provides us with a basis for studying the relationship between evidence and model-based inference, and thereby transparency and control, in settings such as these. Second it provides a practical device that could be implemented by social media sites (and others) to provide such transparency and control to its users.  

%\vspace{-2mm}

\subsection{A Model for Cloaking}
\label{sect:privacycloakabilityandtheevidencecounterfactual:amodelforcloaking}

We will now formally describe cloaking in the context of a linear model (for extension to non-linear models, follow~\cite{MartensProvostMISQ}). Consider predictions of personal traits based on binary Likes as features.
Let $\beta_j$ be the coefficient associated with Like $j \in \{1;...;J\}$. Without loss of generality, assume that these are ranked by decreasing value of $\beta$. Each such coefficient corresponds to the marginal increase in a user's score if he chooses to Like $j$. For a target trait $s$, let $s_i$ be the model output score for user $i$, calculated as $\sum_{j=1}^{J} \beta_j x_{ij}$ where $x_{ij}$ is set to $1$ if user $i$ has Liked $j$ and zero otherwise. We consider ``targeted'' users to be those in the top $\delta$-quantile of the score distribution. Define the cutoff score $s_{\delta}$ to be the score of the highest-ranked user in the quantile directly below the targeted users. Thus the set of targeted users $T_s$ for classification task $s$ is $T_s = \{i \mid s_i > s_{\delta}\}$.

To analyze the difficulty or ease of cloaking for each user in the targeted group, we iteratively remove Likes from his profile in order of the largest effect on the output score.  A user is considered to be successfully cloaked when his score falls below $s_{\delta}$ (and thus he no longer would be targeted). Given a linear model, the Like with the largest effect on the output score is the Like \textit{present in the user's data instance} that has the largest coefficient in the model. For each user and trait the model is static; models are not retrained after each Like is removed. Figure~\ref{fig:removalline} shows two examples.  %\footnote{If the targeted group is defined by a fixed threshold score (such as the estimated probability being above a fixed threshold), this is straightforward.  If the targeted group is defined instead based on the actual quantile, then when a user is removed from the targeted group another user takes his place.  In this paper we consider users in isolation and do not consider the effects of cloaking on sets of users.}
%\footnote{More generally, for non-linear models the evidence counterfactual would reveal a minimal set of Likes such that their removal would successfully cloak the individual~\cite{MartensProvostMISQ}.}
%Figure~\ref{fig:likespace} shows the discriminative power associated with each Like in our data for the task of predicting if individual male users are gay. 
%The ten labeled points depict Likes that have the largest coefficients from the LRSVD model. The top ten Likes for the dashed-diamond user shown in figure~\ref{fig:removalline} are shown here as diamonds. Six out of this user's top-10 Likes overlap with the top ten for the entire task. This highlighted user is the user that the LRSVD model predicts as having the highest probability of being gay.
%To quantify Like removal and the difficulty of cloaking, we let $\eta_{i, \delta}^{s}$ represent the effort to cloak user $i$ from the top $\delta\%$ of the score distribution for a characteristic $s$---specifically, the minimum number of Likes that must be removed to move user $i$ below the threshold. This algorithm is formally described in \textit{SI, Technical Details}.
The absolute effort to cloak
%\footnote{Alternatively, we can examine the relative effort to cloak a task for user $i$, as described in the \textit{Technical Details} section of the \textit{SI}.}
a particular classification target $s$ for user $i$, $\eta_{i, \delta}^{s}$, is given by counting the number of Likes that must be removed for that user to drop below the threshold score. The (average) absolute effort to cloak characteristic $s$ 
is given by averaging across users in $T_s$,

\begin{equation}
\eta_{\delta}^{s} = \frac{\sum_{i \in T_s} \eta_{i, \delta}^{s} } { |T_s| }.
\end{equation}

For the rest of this paper we use $\delta = 0.90$ to indicate that the top $10\%$ of users are being targeted. For other values of $\delta$ the results hold qualitatively.

%\begin{figure}
%  \centerline{\includegraphics[width=1.00\linewidth]{figures/like_space}}
%  \caption{The discriminative power of Likes on Facebook when determining if a user is gay ($Y=1$). Labels are given to the top ten Likes as sorted by their corresponding coefficients from the LRSVD model. Diamond points are the top ten pages Liked by the user with the highest probability of being gay as predicted by the LRSVD model. This is the dash-diamond user that appeared in figure~\ref{fig:removalline}.}
%  \label{fig:likespace}
%\end{figure}

\section{Results}

Let us now examine the effort required to cloak the inferences of a variety of personal characteristics based on data on Facebook users. We first describe the data, and then assess the effort required to cloak user characteristics.

% -------------------------------------------------------

\subsection{Data}
\label{sect:results:data}

Our data were collected through a Facebook application called myPersonality.\footnote{Thanks to the authors of \cite{kosinski2013private} for sharing the data.} Users of the application have opted in and given their consent to have their data recorded for analysis. The data comprise $164,883$ individuals from the United States, including their responses to survey questions and a subset of their Facebook profiles, including Likes. Subsets of users are characterized by their sexual orientation, gender, political affiliation, religious views, IQ, alcohol and drug use, personality dimensions, and lifestyle choices. The personal characteristics are the target variables for the various modeling and inference problems and the Likes are the features. The feature data are very sparse: a user displays less than $0.5\%$ of the set of Likes on average. Table~\ref{table:characteristics} presents summary statistics.%The data are further described in \textit{Data} section of the \textit{SI}.

\begin{table}[h]
\begin{small}
\include{tables/table_data_summary}
\end{small}
\caption{Summary statistics of the dataset. Number of Users indicates how many unique users are associated with the given task. Percent positive shows the percentage of users with true labels for each task. Average Likes indicates the average number of Likes a user associated with the given task has.}
\label{table:characteristics}
\vspace*{-5mm}
\end{table}

We replicated the predictive modeling and inference procedure reported by \cite{kosinski2013private}.  Specifically, we build the predictive models on the top 100 SVD components using logistic regression as implemented in the scikit-learn package in Python (LRSVD). For each model, we choose the regularization parameter by $5$-fold nested cross-validation~\cite{provost2013data}. %Performance of each classification model concurs with those reported by \cite{kosinski2013private} and is outlined in \textit{SI, Classification Performance}. As in the original paper, the predictive performance is quite strong.  

%\vspace{-2mm}

% -------------------------------------------------------

% -------------------------------------------------------

\subsection{Cloaking Difficulty}
\label{sect:results:cloakingdifficulty}

Table~\ref{table:cloakingsummary} reports the effort necessary to cloak users in the top $10\%$ of users as ranked by model score. Column 1 shows that although users display hundreds of Likes, on average they need to cloak less than $10$ of them to successfully inhibit inference. This corresponds to cloaking only about $2-3$\% of a user's Likes, on average. The averages give a fair picture: with only a couple exceptions the proportion of information needed to inhibit inference is around $2-4$\%.  %The actual numbers of Likes that must be removed vary more, as the top-decile users have different total numbers of Likes, but nevertheless there are no extreme outliers.

\begin{figure}
  \centerline{\includegraphics[width=1.00\linewidth]{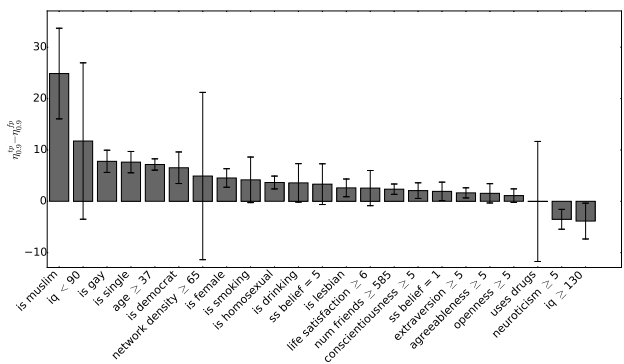}}
  \caption{Difference in cloaking, $\eta_{0.9}$, for true positive and false positive users using the LRSVD model. Error bars depict the 95\% confidence interval.}
  \label{fig:tp-fp:eta}
  \vspace*{-5mm}
\end{figure}

\begin{figure}
  \centerline{\includegraphics[width=1.00\linewidth]{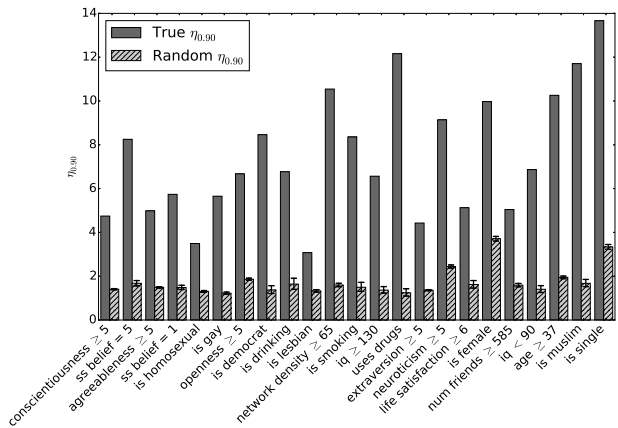}}
  \caption{Comparison between absolute effort ($\eta_{0.9}$) to cloak in the LRSVD model. Results from the normal cloaking procedure are compared to those of a randomization test for each task. Error bars depict the 95\% confidence interval.}
  \label{fig:randomized:eta}
  \vspace*{-5mm}
\end{figure}

To put these results in context it would be useful to know how strongly the cloakability of a trait is related to the statistical dependency structure of the data-generating process. One might think that people who indeed hold a particular trait would exhibit it throughout their behavior, and in particular throughout the things that they Like. How do these cloakability results compare to what one would expect if Likes and the trait were not actually interrelated?
Figure \ref{fig:randomized:eta} shows the difference between $\eta_{0.9}$ in the no-dependency population and the true $\eta_{0.9}$. Additional details on this test can be found in the extended working paper \cite{chen2015inference}.%Across all tasks we find that the actual absolute effort to cloak is higher ($p < 0.01$, sign test) than cloaking would be if Likes were randomly assigned.  Qualitatively, we see that indeed cloaking seems very easy in the random case.  In all but three cases, one needs to cloak fewer than two Likes on average to inhibit inference.  In all cases, inference can be inhibited by cloaking fewer than four Likes on average.  The figure shows that generally the statistical dependency structure renders cloaking several times harder than it would otherwise be.

Although it is indeed relatively easy to inhibit inference by cloaking Likes, the statistical dependencies among the Likes and the predicted trait makes it more difficult than it would be without such dependencies.

\subsection{Cloaking True Positives vs.~False Positives}
\label{sect:results:cloakingtruepositivesvsfalsepositives}

There are multiple settings where one might want to inhibit inference.  Possibly the most important distinction is between inhibiting an inference that is in fact true (a true positive inference) and inhibiting an inference that is false (a false positive inference).  Based on the prior results, one might expect that a false positive inference would be easier to cloak because the statistical dependency to the (positive) trait is by definition missing.  The false positive user was ``accidentally" classified as positive, similarly to how the top-decile randomized users ``accidentally" were classified.  In neither case was the presence of the trait reflected in the behavior of the user. For a more detailed discussion see the working paper \cite{chen2015inference}. %However, there is an important distinction: in the randomized setting the statistical dependencies also were broken among the Likes.  For false positives, intuitively there still may be strong statistical interdependencies between the Likes---if one has some Likes that trigger the inference by the predictive model, one may have many Likes that trigger the inference.

\begin{table}[h]
\caption{The effort necessary to cloak different users' characteristics using the logistic regression with the 100-SVD-component logistic regression (LRSVD) model. The first column shows the full set of users; the second column shows only the true positive users, and the third column shows only the false positive users.}
\begin{small}
\include{tables/eta/table_eta_pi_90p}
\end{small}
\label{table:cloakingsummary}
\end{table}

In table~\ref{table:cloakingsummary}, we show that cloaking is indeed generally more difficult for true-positive users than for false-positive ($p < 0.05$, sign test). The differences in cloakability between true-positive and false-positive users are shown in figure~\ref{fig:tp-fp:eta}.

%These results provide intuitive satisfaction. It is relatively easier to ``fix" an incorrect classification than to ``hide" from a correct inference. A striking example of this is for the ``is muslim'' trait.  On average, to inhibit the positive inference for someone who actually is Muslim, $28$ Likes have to be cloaked.  This is almost twice as many as for any other trait.  On the other hand, to inhibit the ``is muslim" classification for a non-Muslim, only $3$ traits need to be cloaked.  

%This suggests a line of future inquiry: does this illustrate a case of a strong dependency between a personal trait and the individual's choice of actions? Or is there some alternative explanation having to do with the subtleties of predictive modeling? 

%A comprehensive analysis is beyond the scope of this paper, however we offer an initial view in the \textit{SI, Statistical Dependency Relationship}.

% -------------------------------------------------------

\vspace{-2mm}
\subsection{Modeling Behavior}
\label{sect:results:modelingbehavior}

Table \ref{table:cloakingsummarymodels} presents the values for our cloaking measure across different models.\footnote{The generalization performance for the NB model is slightly lower than that for the logistic regression models.}  As expected, the cloaking effort required for the LR and LRSVD models are similar.  In contrast, cloaking is indeed substantially more difficult for NB.  Rather than needing to cloak only a half-dozen or so Likes in the LR and LRSVD models, the NB model's users on average have to cloak $50$ Likes. 
\begin{table}
\caption{The effort to cloak different users' characteristics using logistic regression
with 100 SVD components (LRSVD), logistic regression on the raw Likes (LR), and naive Bayes (NB).}
\begin{small}
\include{tables/eta/table_eta_pi_models}
\end{small}
\label{table:cloakingsummarymodels}
\end{table}

Mechanically, we can explain this difference through the assumptions made by each of the predictive models. The main different between the LR/LRSVD models and the NB model is that the NB algorithm treats Likes as if they are conditionally independent of each other given the target (the trait).  When the Likes in fact are highly correlated, this creates a pathology in predictive behavior: the resulting inference model will tend to ``double count'' when users present correlated Likes. This result indicates that a targeter could choose to make cloaking more difficult without imposing any restrictions on their users simply by changing its predictive model choice (possibly incurring a small loss in predictive performance.)

\section{Conclusion}

The results show that the amount of effort users must exert in order to successfully hide themselves is quite small.  Although it is higher than if there were no statistical dependency among the Likes and the personal traits, the users still need only to cloak about a half-dozen of their hundreds of Likes on average to inhibit inference of a personal trait.  Users for whom the inference made is actually wrong have an even easier time cloaking the inference.  However, organizations engaging in such modeling could alter their modeling choices to make cloaking much more difficult. 

These results raise the question of whether organizations would want to give users such transparency and control.  One argument is that a firm should because, in line with other ``fair information practices" it simply is the right thing to do.  As certain firms become indispensable, individuals would pay a heavy social cost simply to opt out.  An alternative argument is that doing so would increase customer satisfaction, and few people would actually exercise their option for control. In 2014, Facebook introduced a feature called ``Why am I seeing this ad?'' which gives users partial transparency into why they are being targeted. Users can also selectively cloak particular categories of ads or advertisers; they can also modify their ``ad preferences'' to hide categories of information from being used for targeting.  Transparency would be greatly enhanced by revealing which data actually led to the provision of particular content or ads, as well as tools to help users to cloak just the right data.

\bibliography{bibliography}
\bibliographystyle{icml2016}

\end{document}

%% file: tables/table_data_summary.tex
\begin{tabular}{l|rrr}
%\toprule
Task &  \# Users & $\%$ Positive &  Avg. Likes \\
%\midrule
age $\geq$ 37              &        145,400 &          0.127 &        216 \\
agreeableness $\geq$ 5     &        136,974 &          0.014 &        218 \\
conscientiousness $\geq$ 5 &        136,974 &          0.018 &        218 \\
extraversion $\geq$ 5      &        136,974 &          0.033 &        218 \\
iq $\geq$ 130              &          4,540 &          0.130 &        186 \\
iq $<$ 90                  &          4,540 &          0.073 &        186 \\
is democrat                &          7,301 &          0.596 &        262 \\
is drinking                &          3,351 &          0.485 &        262 \\
is female                  &        164,285 &          0.616 &        209 \\
is gay                     &         22,383 &          0.046 &        192 \\
is homosexual              &         51,703 &          0.035 &        257 \\
is lesbian                 &         29,320 &          0.027 &        307 \\
is muslim                  &         11,600 &          0.050 &        238 \\
is single                  &        124,863 &          0.535 &        226 \\
is smoking                 &          3,376 &          0.237 &        261 \\
life satisfaction $\geq$ 6 &          5,958 &          0.125 &        252 \\
network density $\geq$ 65  &         32,704 &          0.012 &        214 \\
neuroticism $\geq$ 5       &        136,974 &          0.004 &        218 \\
num friends $\geq$ 585     &         32,704 &          0.140 &        214 \\
openness $\geq$ 5          &        136,974 &          0.043 &        218 \\
ss belief = 1              &         13,900 &          0.178 &        229 \\
ss belief = 5              &         13,900 &          0.079 &        229 \\
uses drugs                 &          2,490 &          0.172 &        264 \\
%\bottomrule
\end{tabular}

%% file: tables/eta/table_eta_pi_90p.tex
\begin{tabular}{l|rrr}
     &             & $\eta_{0.9}$ &  \\
Task &         All & TP & FP \\
%\midrule
is lesbian                 &   3.075 &     5.437 &      2.829 \\
is homosexual              &   3.493 &     6.572 &      2.888 \\
extraversion $\geq$ 5      &   4.428 &     5.944 &      4.300 \\
conscientiousness $\geq$ 5 &   4.746 &     6.746 &      4.670 \\
agreeableness $\geq$ 5     &   4.985 &     6.508 &      4.957 \\
num friends $\geq$ 585     &   5.043 &     6.556 &      4.197 \\
life satisfaction $\geq$ 6 &   5.128 &     7.214 &      4.642 \\
is gay                     &   5.653 &    10.944 &      3.161 \\
ss belief = 1              &   5.738 &     6.880 &      4.946 \\
iq $\geq$ 130              &   6.566 &     3.429 &      7.283 \\
openness $\geq$ 5          &   6.674 &     7.677 &      6.571 \\
is drinking                &   6.771 &     7.463 &      3.875 \\
iq $<$ 90                  &   6.867 &    16.318 &      4.582 \\
ss belief = 5              &   8.251 &    11.098 &      7.760 \\
is smoking                 &   8.357 &     9.800 &      5.621 \\
is democrat                &   8.462 &     8.533 &      2.000 \\
neuroticism $\geq$ 5       &   9.140 &     5.667 &      9.173 \\
is female                  &   9.971 &    10.015 &      5.475 \\
age $\geq$ 37              &  10.259 &    13.011 &      5.847 \\
network density $\geq$ 65  &  10.545 &    15.308 &     10.388 \\
is muslim                  &  11.706 &    27.804 &      2.930 \\
uses drugs                 &  12.161 &    12.143 &     12.176 \\
is single                  &  13.665 &    15.514 &      7.888 \\
\textbf{Mean}              & \textbf{7.465}& \textbf{9.851}& \textbf{5.572} \\
% \textbf{Median}            & \textbf{6.771}& \textbf{7.677}& \textbf{4.946} \\
%\bottomrule
\end{tabular}

%% file: tables/eta/table_eta_pi_models.tex
\begin{tabular}{l|rrr}
                            &                & $\eta_{0.9}$  &                 \\
Task                        & LRSVD          & LR            & NB              \\
is lesbian                  & 3.075          & 2.507         & 15.518     \\
is homosexual               & 3.493          & 3.396         & 17.046     \\
extraversion $\geq$ 5       & 4.428          & 3.617         & 58.048    \\
conscientiousness $\geq$ 5  & 4.746          & 3.357         & 24.471    \\
agreeableness $\geq$ 5      & 4.985          & 2.871         & 8.727     \\
num friends $\geq$ 585      & 5.043          & 4.748         & 57.596    \\
life satisfaction $\geq$ 6  & 5.128          & 4.061         & 11.959    \\
is gay                      & 5.653          & 9.073         & 17.796    \\
ss belief = 1               & 5.738          & 4.550         & 19.908    \\
iq $\geq$ 130               & 6.566          & 2.920         & 21.947    \\
openness $\geq$ 5           & 6.674          & 3.700         & 37.433    \\
is drinking                 & 6.771          & 5.398         & 17.687    \\
iq $<$ 90                   & 6.867          & 3.681         & 7.664    \\
ss belief = 5               & 8.251          & 4.692         & 25.450    \\
is smoking                  & 8.357          & 7.012         & 44.821    \\
is democrat                 & 8.462          & 9.396         & 65.363    \\
neuroticism $\geq$ 5        & 9.140          & 2.292         & 125.232   \\
is female                   & 9.971          & 11.619        & 320.665    \\
age $\geq$ 37               & 10.259         & 7.263         & 37.746    \\
network density $\geq$ 65   & 10.545         & 2.569         & 75.717    \\
is muslim                   & 11.706         & 8.934         & 31.131    \\
uses drugs                  & 12.161         & 8.161         & 20.532    \\
is single                   & 13.665         & 10.233        & 101.658   \\
\textbf{Mean}               & \textbf{7.465} & \textbf{5.48} & \textbf{50.614} \\
\end{tabular}